\documentclass[11pt,a4paper]{article}
\usepackage[hyperref]{acl2020}
\usepackage{times}
\usepackage{latexsym}

\usepackage[utf8]{inputenc} 
\usepackage[T1]{fontenc}    
\usepackage{hyperref}       
\usepackage{url}            
\usepackage{booktabs}       
\usepackage{amsfonts}       
\usepackage{nicefrac}       
\usepackage{microtype}      
\usepackage{xspace}
\usepackage{amssymb}
\usepackage{mathrsfs}
\usepackage{algorithm}
\usepackage{algorithmic}
\usepackage{xcolor}
\usepackage{enumitem}
\usepackage{haldefs}
\usepackage{graphicx}

\usepackage{microtype}

\aclfinalcopy 

\title{Meta-Learning for Few-Shot NMT Adaptation}

\author{Amr Sharaf \\
  University of Maryland \\
  \texttt{amr@cs.umd.edu} \\\And
  Hany Hassan \\
  Microsoft \\
  \texttt{hanyh@microsoft.com} \\\And
  Hal Daumé III \\
  Microsoft Research \&\\
  University of Maryland \\
  \texttt{me@hal3.name}}

\date{}

\newcommand{\ourname}{\textsc{Meta-MT}\xspace}
\newcommand{\ournamelong}{Meta-learning for Machine Translation\xspace}
\newcommand\Tau{\mathrm{T}}
\newcommand\TTau{\mathcal{T}}
\newcommand\p{\mathrm{P}}
\newcommand{\algorithmname}[1]{\textsc{#1}\xspace}

\def\equationautorefname~#1\null{Eq~#1\null}
\renewcommand{\sectionautorefname}{\S\kern-0.2em}
\renewcommand{\subsectionautorefname}{\S\kern-0.2em}
\renewcommand{\subsubsectionautorefname}{\S\kern-0.2em}

\begin{document}
\maketitle
\begin{abstract}
We present \ourname, a meta-learning approach to adapt Neural Machine Translation (NMT) systems in a few-shot setting. \ourname provides a new approach to make NMT models  easily adaptable to many target  domains with the minimal  amount of in-domain  data. We frame the adaptation of NMT systems as a meta-learning problem, where we learn to adapt to new unseen domains based on simulated offline meta-training domain adaptation tasks. We evaluate the proposed meta-learning strategy on ten domains with general large scale NMT systems. We show that \ourname significantly outperforms classical domain adaptation when very few in-domain  examples are available. Our experiments shows that \ourname can outperform classical fine-tuning by up to  2.5 BLEU points after seeing only $4,000$ translated words ($300$ parallel sentences). 
\end{abstract}

\section{Introduction}
Neural Machine Translation (NMT) systems~\citep{Bahdanau2016,NIPS2014_5346} are usually trained on large general-domain
parallel corpora to achieve state-of-the-art results~\citep{barrault2019findings}. Unfortunately, these generic corpora
are often qualitatively different from the target domain of the translation system. Moreover, NMT models trained on one
domain tend to perform poorly when translating sentences in a significantly different
domain~\citep{koehn-knowles-2017-six,chu-wang-2018-survey}. A widely used approach for adapting NMT is \emph{domain
adaptation by fine-tuning}~\citep{Luong-Manning:iwslt15,Freitag2016FastDA,sennrich-etal-2016-improving}, where a model
is first trained on general-domain data and then adapted by continuing the training on a smaller amount of in-domain
data. This approach often  leads to empirical improvements in the targeted domain; however, it  falls short when the
amount of in-domain training data is insufficient, leading to model over-fitting and catastrophic forgetting, where
adapting to a new domain leads to degradation on the general-domain~\citep{thompson-etal-2019-overcoming}.
Ideally, we would like to have a model that is easily adaptable to many target domains with  minimal  amount of in-domain  data.

We present  a meta-learning approach that \emph{learns} to adapt neural machine translation systems to new domains given
only a small amount of training data in that domain. To achieve this, we simulate many domain adaptation tasks, on which
we use a \emph{meta-learning} strategy to learn how to adapt. Specifically, based on these simulations, our proposed
approach, \ourname (\ournamelong), learns model parameters that should generalize to future (real) adaptation tasks
(\autoref{sec:test}).

At training time (\autoref{sec:train}), \ourname simulates many small-data domain adaptation tasks from a large pool of
data. Using these tasks, \ourname simulates what would happen after fine-tuning the model parameters to each such task.
It then uses this information to compute parameter updates that will lead to efficient adaptation during deployment. We
optimize this using the Model Agnostic Meta-Learning algorithm (MAML)~\citep{pmlr-v70-finn17a}. 

The contribution of this paper is as follows: first, we propose a new approach that enables NMT systems to effectively
adapt to a new domain using few-shots learning. Second, we show what models and conditions enable meta-learning to be
useful  for NMT adaptation. Finally,  We evaluate \ourname on ten different domains, showing the efficacy of our
approach. To the best of our knowledge, this is the first work on  adapting large scale NMT systems in a few-shot
learning setup~\footnote{\textbf{Code Release:} We make the code publicly available online: \url{https://www.dropbox.com/s/jguxb75utg1dmxl/meta-mt.zip?dl=0}}.

\section{Related Work}

Our goal for  few-shot NMT adaptation is to adapt a pre-trained NMT model (e.g. trained on general domain data) to new
domains (e.g. medical domain) with a small amount of training examples. \citet{chu2018comprehensive} surveyed several
recent approaches  that address the shortcomings of traditional fine-tuning when applied to domain adaptation. Our work
distinguishes itself from prior work by learning to fine-tune with tiny amounts of training examples.  

Most recently, \citet{bapna2019simple} proposed a simple approach for adaptation in NMT. The approach consists of
injecting task specific adapter layers into a pre-trained model. These adapters enable the model to adapt to new tasks
as it introduces a bottleneck in the architecture that makes it easier to adapt. Our approach uses a similar model
architecture, however, instead of injecting a new adapter for each task separately, \ourname uses a single adapter
layer, and meta-learns a better initialization for this layer that can easily be fine-tuned to new domains with very few
training  examples.

Similar to our goal, \citet{michelneubig2018extreme} proposed a space efficient approach to adaptation that learns
domain specific biases to the output vocabulary. This enables large-scale personalization for NMT models when small
amounts of data are available for a lot of different domains. However, this approach assumes that these domains are
static and known at training time, while \ourname can dynamically generalize to totally new domains, previously unseen
at meta-training time.   

Several approaches have been proposed for lightweight adaptation of NMT systems.~\citet{vilar-2018-learning} introduced
domain specific gates to control the contribution of hidden units feeding into the next layer. However,
\citet{bapna2019simple} showed that this introduced a limited amount of per-domain capacity; in addition, the learned
gates are not guaranteed to be easily adaptable to unseen domains.~\citet{khayrallah-etal-2017-neural} proposed a
lattice search algorithm for NMT adaptation, however, this algorithm assumes access to lattices generated from a phrase
based machine translation system. 

Our meta-learning strategy mirrors that of \citet{gu-etal-2018-meta} in the low resource translation setting, as well as
\citet{wu2019enhanced} for cross-lingual named entity recognition with minimal resources,
\citet{Mi:2019:MLN:3367471.3367479} for low-resource natural language generation in task-oriented dialogue systems, and
\citet{dou-etal-2019-investigating} for low-resource natural language understanding tasks. To the best of our knowledge,
this is the first work using  meta-learning for few-shot NMT adaptation.

\begin{toappendix}
\section{Background}
\subsection{Neural Machine Translation}
Neural Machine Translation (NMT) is a sequence to sequence model that parametrizes  the conditional probability of the source and target sequences as a neural network following encoder-decoder architecture~\citep{Bahdanau2016,NIPS2014_5346}. Initially, the encode-decoder architecture was represented by recurrent networks. Currently, this has been replaced by self-attention models aka Transformer models~\citep{vaswani2017attention}). 
Currently, Transformer models achieves state-of-the-art performance in NMT as well as many other language modeling tasks. While transformers models are performing quite well on large scale NMT tasks, the models have huge number of parameters  and require large
amount of training data which is really prohibitive for
adaptation tasks especially in few-shot setup like ours.

\subsection {Few Shots Domain Adaptation}

Traditional domain adaptation for NMT models assumes the availability of  relatively large amount of in domain data. For
instances most of the related work utilizing traditional fine-tuning experiment with hundred-thousand sentences
in-domain. This setup in quite prohibitive, since practically the domain can be defined by few examples. In this work we
focus on few-shot adaptation scenario where we can adapt to a new domain not seen during training  time using  just
couple of hundreds  of in-domain sentences. This introduces  a new challenge where the models have to be quickly
responsive to adaptation as well as robust to domain shift. Since we focus on the setting in which very few in-domain data is available, this renders many  traditional domain adaptation approaches inappropriate.

\subsection{Meta-Learning}

Meta-learning or Learn-to-Learn is widely used for few-shot learning in many applications where a model trained for a particular task can  learn  another task with a few examples. A number of approaches are used in Meta-learning, namely:  Model-agnostic Meta-Learning (MAML) and its first order approximations like First-order MAML (FoMAML)~\citep{pmlr-v70-finn17a} and Reptile~\citep{nichol2018first}. In this work, we focus on using MAML to enable few-shots adaptation of NMT transformer models.
\end{toappendix}

\section{Approach: Meta-Learning for Few-Shot NMT Adaptation}

Neural Machine Translation systems are not robust to domain shifts~\citep{chu-wang-2018-survey}. It is a highly
desirable characteristic  of the system to be  adaptive to any domain shift using  weak supervision without degrading
the performance  on  the general domain. This  dynamic adaptation task can be viewed  naturally as a learning-to-learn
(meta-learning) problem: how can we train a global model that is capable of using its previous experience in adaptation
to learn to adapt faster to unseen domains? A particularly simple and effective strategy for adaptation is fine-tuning:
the global model is adapted by training on in-domain data.  One would hope to improve on such a strategy by decreasing
the amount of required in-domain data.  \ourname takes into account information from previous adaptation tasks, and aims
at learning how to update the global model parameters, so that the resulting learned parameters after meta-learning can
be adapted faster and better to previously unseen domains via a weakly supervised fine-tuning approach on a tiny amount
of data. 

Our goal in this paper is to learn how to adapt a neural machine translation system from experience.  The training
procedure for \ourname  uses offline simulated adaptation problems to learn model parameters $\theta$ which can adapt
faster to previously unseen domains.  In this section, we describe \ourname, first by describing how it operates at test
time when applied to a new domain adaptation task (\autoref{sec:test}), and then by describing how to train it using
offline simulated adaptation tasks (\autoref{sec:train}). 

\begin{figure}
    \centering
    \includegraphics[width=0.5\textwidth]{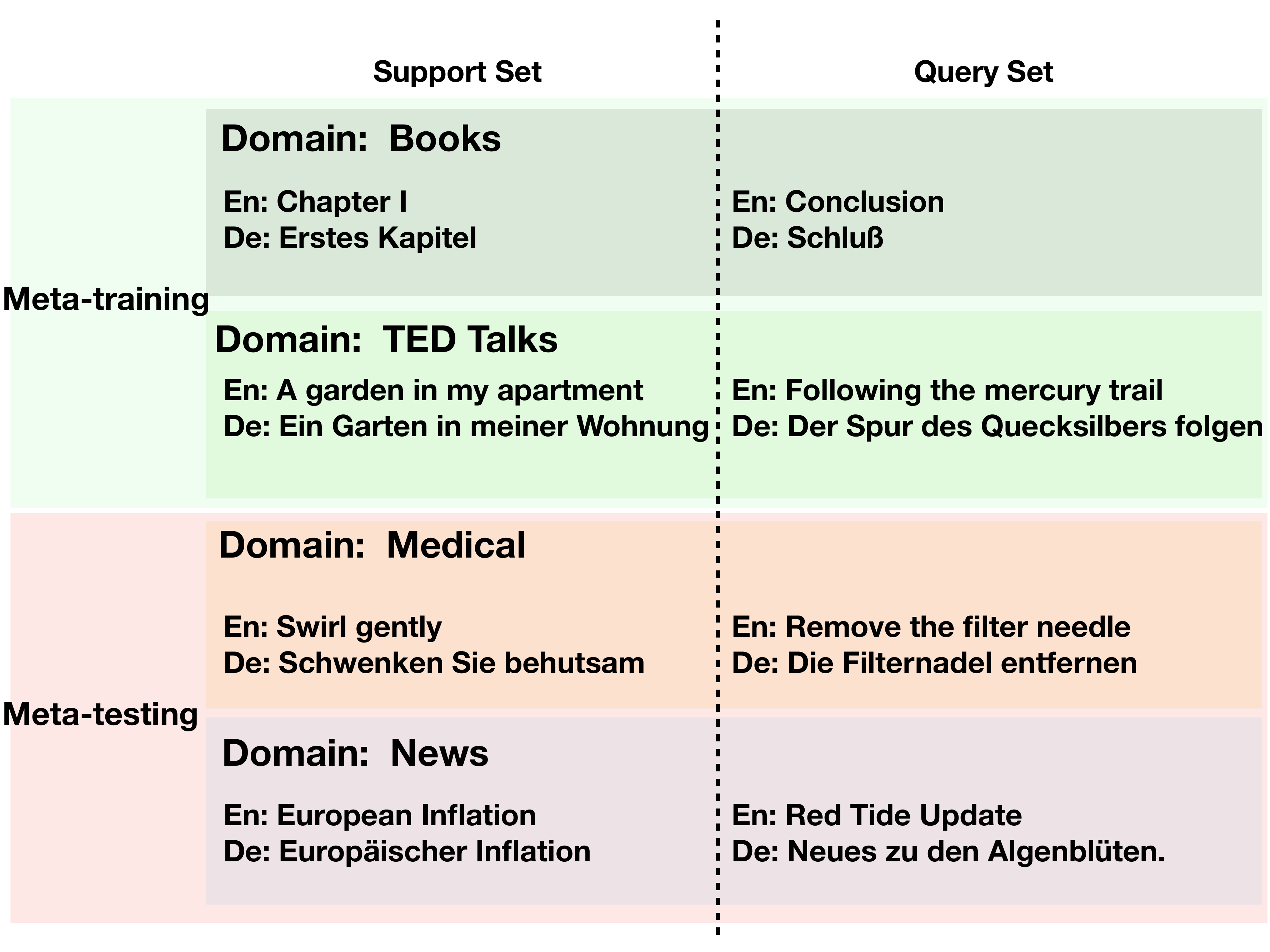}
    \caption{
 Example meta-learning set-up for few-shot NMT adaptation. The top represents the meta-training set
    $\cal D_{\textrm{meta-train}}$, where inside each box is a separate dataset $\Tau$ that consists of the support set
    $\Tau_{\textrm{support}}$ (left side of dashed line) and the query set $\Tau_{\textrm{query}}$ (right side of dashed 
    line). In this illustration, we are considering the books and TED talks domains for meta-training. The meta-test set 
    $\cal D_{\textrm{meta-test}}$ is defined in the same way, but with a different set of domains  not present in any of 
    the datasets in $\cal D_{\textrm{meta-train}}$: Medical and News.
    }
    \label{fig:main}
\end{figure}

\begin{figure}
    \centering
    \includegraphics[width=0.48\textwidth]{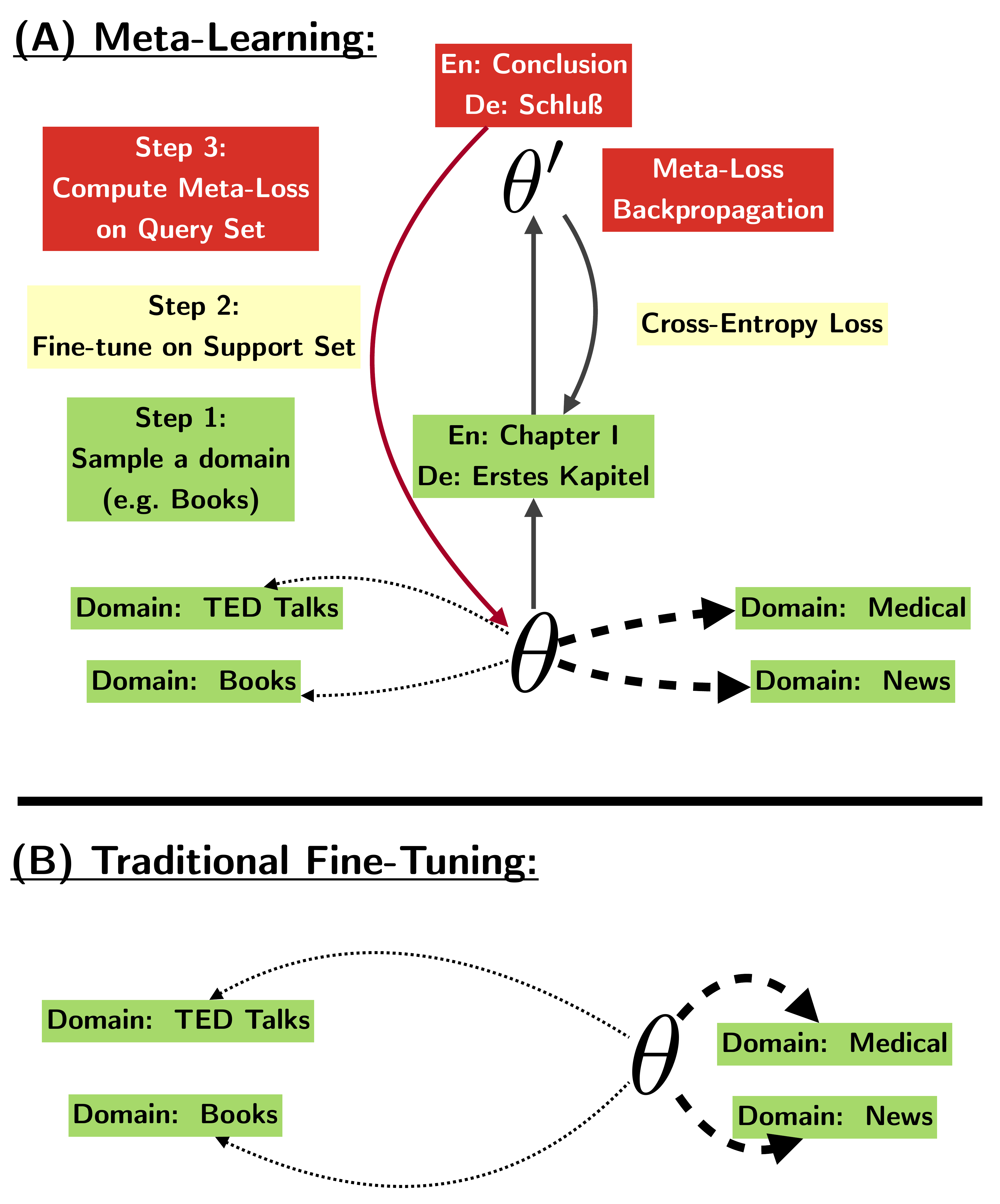}
    \caption{\textbf{[Top-A]} a training step of \ourname. \textbf{[Bottom-B]} Differences between  meta-learning and Traditional fine-tuning. 
    Wide lines represent high resource domains (Medical, News), while thin lines represent low-resource domains (TED, Books). 
    Traditional fine-tuning may favor high-resource domains over low-resource ones while meta-learning aims at 
    learning a good initialization that can be adapted to any domain with minimal training samples. \footnotemark}
    \label{fig:main:meta}
\end{figure}

\subsection{Test Time Behavior of \ourname}
\label{sec:test}

  At test time, \ourname adapts a pre-trained NMT model to a new given domain. The adaptation is done using a small
  in-domain data that we call the \emph{support set}  and then tested on the new domain using a \emph{query set}. More
  formally, the model parametrized by $\theta$ takes as input a new adaptation task $\Tau$. This is illustrated
  in~\autoref{fig:main}: the adaptation task $\Tau$ consists of a standard domain adaptation problem: $\Tau$ includes a
  support set $\Tau_{\textrm{support}}$ used for training the fine-tuned model, and a query set $\Tau_{\textrm{query}}$
  used for evaluation. We're particularly interested in the distribution of tasks $\p(\TTau)$ where the support and
  query sets are very small. In our experiments, we restrict the size of these sets to only few hundred parallel
  training sentences. We consider support sets of sizes: 4k to 64k source words (i.e. $\sim 200$ to $3200$ sentences).
  At test time, the meta-learned model $\theta$ interacts with the world as follows (\autoref{fig:main:meta}):
  \begin{enumerate}[noitemsep,nolistsep] 
  \item \textbf{Step 1:} The world draws an adaptation task $\Tau$ from a distribution $\p$, $\Tau
  \sim \p(\TTau)$; 
  \item \textbf{Step 2:} The model adapts from $\theta$ to $\theta'$ by fine-tuning on the task's support set
  $\Tau_{\textrm{support}}$; 
  \item \textbf{Step 3:} The fine-tuned model $\theta'$ is evaluated on the query set $\Tau_{\textrm{query}}$.
  \end{enumerate}

Intuitively, meta-training should optimize for a representation $\theta$ that can quickly adapt to new tasks, rather than a single individual task. 

\subsection{Training \ourname via Meta-learning}
\label{sec:train}

\footnotetext{colorblind friendly palette was selected from~\citet{neuwirth2014colorbrewer}.}

The meta-learning challenge is: how do we learn a good representation $\theta$?  We initialize $\theta$ by training an
NMT model on global-domain data. In addition, we assume access to meta-training tasks on which we can train $\theta$;
these tasks must include support/query pairs, where we can simulate a domain adaptation setting by fine-tuning on the
support set and then evaluating on the query. This is a weak assumption: in practice, we use purely simulated data as
this meta-training data. We construct this data as follows: given a parallel corpus for the desired language pair, we
randomly sample training example to form a few-shot adaptation task. We build tasks of 4k, 8k, 16k, 32k, and 64k
training words. Under this formulation, it's natural to think of $\theta$'s learning process as a process to learn a
good parameter initialization for fast adaptation, for which a class of learning algorithms to consider are
Model-agnostic Meta-Learning (MAML) and it's first order approximations like First-order MAML
(FoMAML)~\citep{pmlr-v70-finn17a} and Reptile~\citep{nichol2018first}. 

Informally, at training time, \ourname will treat one of these simulated domains $\Tau$ as if it were a domain
adaptation dataset. At each time step, it will update the current model representation from $\theta$ to $\theta'$ by
fine-tuning on $\Tau_{\textrm{support}}$ and then ask: what is  the meta-learning loss estimate given $\theta$,
$\theta'$, and $\Tau_{query}$? The model representation $\theta$ is then updated to minimize this meta-learning loss.
More formally, in meta-learning, we assume access to a distribution $\p$ over different tasks $\TTau$. From this, we can
sample a meta-training dataset $\cal D_{\textrm{meta-train}}$. The meta-learning problem is then to estimate $\theta$ to
minimize the meta-learning loss on $\cal D_{\textrm{meta-train}}$.

\begin{algorithm}[t]
  \caption{\algorithmname{\ourname}(trained model $f_{\theta}$, meta-training dataset $\cal D_{\textrm{meta-train}}$, learning rates $\alpha, \beta$)}
  \label{alg:train}
    \begin{algorithmic}[1]
        \WHILE{not done} 
            \STATE{Sample a batch of domain adaptation tasks $\Tau \sim \cal D_{\textrm{meta-train}}$}\label{train:sample}
             \FORALL{$\Tau_i \in \Tau$} 
                \STATE{Evaluate $\nabla_{\theta}L_{\Tau_i}(f_{\theta})$ on the support set $\Tau_{i,\textrm{support}}$}\label{train:adapt} 
                \STATE{Compute adapted parameters with gradient descent: $\theta'_{i}= \theta - \alpha \nabla_{\theta} L_{\Tau_i}(f_{\theta})$}\label{train:gradient}
             \ENDFOR
             \STATE{Update $\theta \leftarrow \theta - \beta \nabla_{\theta}\Sigma_{\Tau_i \in \Tau} L_{\Tau_i}(f_{\theta'_{i}})$ on the query set $\Tau_{i,\textrm{query}} \forall \Tau_i \in \Tau$}\label{train:loss} 
        \ENDWHILE
  \end{algorithmic}
\end{algorithm}

The meta-learning algorithm we use is MAML by~\citet{pmlr-v70-finn17a}, and is instantiated for the meta-learning to adapt NMT systems in~\autoref{alg:train}. MAML considers a model represented by a parametrized function $f_{\theta}$ with parameters $\theta$. When adapting to a new task $\Tau$, the model’s parameters $\theta$ become $\theta'$. The updated  vector $\theta'$ is computed
using one or more gradient descent updates on the task $\Tau$. For example, when using one gradient update:

\begin{align}
\theta' = \theta - \alpha \nabla_{\theta}L_{\Tau}(f_{\theta})
\end{align}

where $\alpha$ is the learning rate and $L$ is the task loss function. The model parameters are trained by optimizing for the performance of $f_{\theta'}$ with respect to $\theta$ across tasks sampled from $\p(\TTau)$. More concretely, the meta-learning objective is:
\begin{align}
\notag
\min_{\theta} \Sigma_{\Tau \sim \p(\TTau)}  L_{\Tau} (f_{\theta'}), \\
L_{\Tau} (f_{\theta'}) = L_{\Tau} (f_{\theta - \alpha \nabla_{\theta}L_{\Tau}(f_{\theta})})
\label{eq:loss}
\end{align}

Following the MAML template, \ourname operates in an iterative fashion, starting with a trained NMT model $f_{\theta}$ and improving it through optimizing the meta-learning loss from~\autoref{eq:loss} on the meta-training dataset $\cal D_{\textrm{meta-train}}$. Over learning rounds, \ourname selects a random batch of training tasks from the meta-training dataset and simulates the test-time behavior on these tasks (Line \autoref{train:sample}). The core functionality is to observe how the current model representation $\theta$ is adapted for each task in the batch, and to use this information to improve $\theta$ by optimizing the meta-learning loss (Line \autoref{train:loss}). \ourname achieves this by simulating a domain adaptation setting by fine-tuning on the task specific support set (Line \autoref{train:adapt}). This yields, for each task $\Tau_i$, a new adapted set of parameters $\theta'_i$ (Line \autoref{train:gradient}).
These parameters are evaluated on the query sets for each task $\Tau_{i,\textrm{query}}$, and a meta-gradient w.r.t the original model representation $\theta$ is used to improve $\theta$ (Line \autoref{train:loss}).

Our  pre-trained baseline NMT model $f_{\theta}$  is  a sequence to sequence model that parametrizes  the conditional probability of the source and target sequences as an  encoder-decoder architecture using self-attention  Transformer models~\citep{vaswani2017attention}).

\section{Experimental Setup and Results}
\label{sec:data}
 We seek to answer the following questions experimentally: 

\begin{enumerate}[noitemsep,nolistsep]
    \item How does \ourname compare empirically to alternative adaptation strategies? (\autoref{sec:results})
    \item What is the impact of the support and the query sizes used for meta-learning? (\autoref{sec:size})
    \item What is the effect of the NMT model architecture on performance?
    (\autoref{sec:architecture})
\end{enumerate}

\begin{toappendix}
\section{Statistics of in-domain data sets}
\begin{table}
\centering
\begin{tabular}{lrl}
\toprule 
\textbf{Domain} & \textbf{\# sentences} & \textbf{\# En Tokens} \\ 
\midrule
bible-uedin & 62195 & 1550431\\
ECB & 113174 & 3061513\\
KDE4 & 224035 & 1746216\\
Tanzil & 537128 & 9489824\\
WMT-News & 912212 & 5462820\\
Books & 51467 & 1054718\\
EMEA & 1108752 & 12322425\\
GlobalVoices & 66650 & 1239921\\
ufal-Med & 140600 & 5527010\\
TED & 51368 & 1060765\\
\bottomrule
\end{tabular}
\caption{\label{tab:stats} Dataset statistics for different domains.}
\end{table}

Table~\ref{tab:stats} lists the sizes of various in-domain datasets from which we sample our in-domain data to simulate the few-shot adaptation setup.

\end{toappendix}

In our experiments, we train \ourname only on simulated data, where we simulate a few-shot domain adaptation setting as described
in~\autoref{sec:train}. This is possible because \ourname learns model parameters $\theta$ that can generalize to future
adaptation tasks by optimizing the meta-objective function in~\autoref{eq:loss}.

We train and evaluate \ourname on a collection of ten different datasets. All of these datasets are collected from the
Open Parallel Corpus (OPUS)~\cite{TIEDEMANN12463}, and are publicly available online. The datasets cover a variety
of diverse domains that should enable us to evaluate our proposed approach. The datasets we consider are: 

\begin{enumerate}[noitemsep,nolistsep]
    \item Bible: a parallel corpus created from translations of the
    Bible~\cite{christodouloupoulos2015massively}. 
    \item European Central Bank: website and documentations from the European Central Bank.
    \item KDE: a corpus of KDE4 localization files.
    \item Quran: a collection of Quran translations compiled by the Tanzil project.
    \item WMT news test sets: a parallel corpus of News Test Sets provided by WMT.
    \item Books: a collection of copyright free books.
    \item European Medicines Agency (EMEA): a parallel corpus made out of PDF documents from the European Medicines Agency. 
    \item Global Voices: parallel news stories from the Global Voices web site.
    \item Medical (ufal-Med): the UFAL medical domain dataset from~\citet{yepes2017findings}.
    \item TED talks: talk subtitles from ~\citet{duh18multitarget}.
\end{enumerate}

We simulate the few-shot NMT adaptation scenarios by randomly sub-sampling these datasets with different sizes. We
sample different data sets with sizes ranging from 4k to 64k training words (i.e. $\sim 200$ to $3200$ sentences). This
data is the only data used for any given domain across all adaptation setups. It is worth noting that different datasets
have  a wide range of sentence lengths. We opted to sample using number of words instead of number of sentences to avoid
introducing any advantages for domains with longer sentences. 

\subsection{Domain Adaptation Approaches}
\label{sec:baselines}

Our experiments aim to determine how \ourname compares to standard domain adaptation strategies. In particular, we compare to:
\begin{enumerate}[label=(\Alph*),noitemsep,nolistsep]
    \item \textbf{No fine-tuning:} \label{item:a} The non-adaptive baseline. Here, the pre-trained model is evaluated on
    the meta-test and meta-validation datasets (see \autoref{fig:main}) without any kind of adaptation. 
    \item \textbf{Fine-tuning on a single task:} \label{item:b} The domain adaptation by fine-tuning  baseline. For a
    single adaptation task $\Tau$, this approach performs domain adaptation by fine-tuning only on the support set
    $\Tau_{\textrm{support}}$. For instance, if $|\Tau_{\textrm{support}}| = K$ words, we fine tune the pre-trained
    model $f_{\theta} $only on $K$ training words to show how classical fine-tuning behaves in few-shot settings.
    \item \textbf{Fine-tuning on meta-train:} \label{item:c} Similar to~\autoref{item:b}, however, this approach
    fine-tunes on much more data. This approach fine-tunes on all the support sets used for meta-training:
    $\{\Tau_{\textrm{support}}, \forall \Tau \in \cal D_{\textrm{meta-train}}\}$. The goal of this baseline is to ensure
    that \ourname doesn't get an additional advantage by training on more data during the meta-training phase.  For
    instance,  if we are using $N$ adaptation tasks each with a support set of size $K$, this will be using $N*K$ words
    for classical fine-tuning. This establishes a fair baseline to evaluate how classical fine-tuning would perform using the same data albeit in a different configuration.
    \item \textbf{\ourname:} \label{item:e} Our proposed approach from~\autoref{alg:train}.  In this setup, we use
    $N$ adaptation tasks $\Tau$ in $\cal D_{\textrm{meta-train}}$,  each with a support set of size $K$ words to perform Meta-Learning. Second order meta-gradients are ignored to decrease the computational complexity.
\end{enumerate}

\subsection{Model Architecture and Implementation Details}

We use the Transformer Model~\citep{vaswani2017attention} implemented in  fairseq~\citep{ott2019fairseq}. In this work,
we use a transformer model with a modified architecture that can facilitate better adaptation. We use \emph{``Adapter
Modules''}~\citep{Houlsby-adapters,bapna2019simple} which introduce an extra layer after each transformer block that can
enable more efficient tuning of the models.  Following~\citet{bapna2019simple}, we augment the Transformer model with
feed-forward adapters: simple single hidden-layer feed-forward networks, with a nonlinear activation function between
the two projection layers. These adapter modules are introduced after the Layer Norm and before the residual connection
layers.  It is composed of a down  projection layer, followed by  a ReLU,  followed by an up projection layer.  This
bottle-necked module with fewer parameters  is very attractive for domain adaptation as we will discuss
in~\autoref{sec:architecture}.  These modules are introduced after every layer in both the encoder and the decoder.  All
experiments are based on the ``base'' transformer model with six blocks in the encoder and decoder networks. Each
encoder block contains a self-attention layer, followed by two fully connected feed-forward layers with a ReLU
non-linearity between them. Each decoder block contains self-attention, followed by encoder-decoder attention, followed
by two fully connected feed-forward layers with a ReLU non-linearity between them.

We use word representations of size $512$, feed-forward layers with inner dimensions $2,048$, multi-head attention with
$8$ attention heads, and adapter modules with $32$ hidden units. We apply dropout~\citep{srivastava2014dropout} with
probability $0.1$.  The model is optimized with Adam~\citep{kingma2014adam} using $\beta_1=0.9, \beta_2=0.98$, and a
learning rate $\alpha=7e-4$. We use the same learning rate schedule as \citet{vaswani2017attention} where the learning
rate increases linearly for $4,000$ steps to $7e-4$, after which it is decayed proportionally to the inverse square root
of the number of steps. For meta-learning, we used a meta-batch size of $1$. We optimized the meta-learning loss
function using Adam with a learning rate of $1e-5$ and default parameters for $\beta_1, \beta_2$. 

All data is pre-processed with joint sentence-pieces \citep{kudo-richardson-2018-sentencepiece} of size 40k. In all
cases, the baseline machine translation system is a neural English to German (En-De) transformer
model~\citep{vaswani2017attention}, initially trained on 5.2M sentences filtered from the the standard parallel  data
(Europarl-v9, CommonCrawl, NewsCommentary-v14, wikititles-v1 and Rapid-2019) from the WMT-19 shared
task~\citep{barrault2019findings}.  We use WMT14  and WMT19 newtests  as validation  and test sets  respectively. The
baseline system scores 37.99 BLEU on the full WMT19 newstest which  compares favorably  with strong single system
baselines at WMT19 shared task~\citep{ng-etal-2019-facebook,JunczysDowmunt2019MicrosoftTA}.

For meta-learning, we use the MAML algorithm as described in~\autoref{alg:train}. To minimize memory consumption, we
ignored the second order gradient terms from~\autoref{eq:loss}. We implement the First-Order MAML approximation (FoMAML)
as described in~\citet{pmlr-v70-finn17a}. We also experimented with the first-order meta-learning algorithm
Reptile~\cite{nichol2018first}. We found that since Reptile doesn't directly account for the performance on the task
query set, along with the large model capacity of  the Transformer architecture, it can easily over-fit to the support
set, thus achieving almost perfect performance on the support, while the performance on the query degrades
significantly.  Even after performing early stopping on the query set, Reptile didn't account correctly for learning
rate scheduling, and finding suitable learning rates for optimizing the meta-learner and the task adaptation was
difficult.  In our experiments, we found it essential to match the behavior of the dropout layers when computing the
meta-objective function in~\autoref{eq:loss} with the test-time behavior described in~\autoref{sec:test}. In particular,
the model has to run in \emph{``evaluation mode''} when computing the loss on the task query set to match the test-time
behavior during evaluation.

\begin{table*}[t]
  \centering
  \begin{tabular}{ccccc}
    \toprule
    Domain     & \textsf{\tiny\textcolor{gray}{A.}} No fine-tuning     & \textsf{\tiny\textcolor{gray}{B.}} Fine-tuning
    on task & \textsf{\tiny\textcolor{gray}{C.}} Fine-tuning on meta-train & \textsf{\tiny\textcolor{gray}{D.}}
    \ourname \\
    \midrule
    Books & $11.338 \pm 0.25$ & $11.34 \pm 0.24$ & $\underline{12.49 \pm 0.15}$ & $\mathbf{12.92 \pm 0.94}$\\
    Tanzil & $11.25 \pm 0.04$ & $11.33 \pm 0.04$ & $\underline{13.62 \pm 0.05}$ & $\mathbf{15.16 \pm 0.94}$\\
    Bible & $12.93 \pm 0.93$ & $12.95 \pm 0.94$ & $17.19 \pm 0.54$ & $\mathbf{24.70 \pm 0.61}$\\
    KDE4 & $20.53 \pm 0.34$ &	$20.54 \pm 0.32$ & $\underline{26.61 \pm 0.16}$ &	$\mathbf{27.26 \pm 0.36}$\\
    Med & $19.30 \pm 0.24$ & $19.53 \pm 0.28$ & $28.31 \pm 0.04$ &	$\mathbf{29.59 \pm 0.05}$\\
    GlobalVoices & $25.10 \pm 0.11$ &	$25.17 \pm 0.23$ & $25.83 \pm 0.25$ &	$\mathbf{26.03 \pm 0.13}$\\
    WMT-News & $26.93 \pm 0.36$ &	$26.92 \pm 0.48$ & $\mathbf{27.26 \pm 0.55}$ &	$\underline{27.23 \pm 0.12}$\\
    TED & $27.69 \pm 0.05$ &	$27.85 \pm 0.06$ & $28.78 \pm	0.03$ & $\mathbf{29.37 \pm 0.03}$\\
    EMEA & $27.81 \pm 0.01$ &	$27.79 \pm 0.05$ & $29.77 \pm 0.59$ & $\mathbf{32.38 \pm 0.01}$\\
    ECB & $29.18 \pm 0.03$ & $29.21 \pm 0.04$ &	$31.18 \pm 0.01$ & $\mathbf{33.23 \pm 0.40}$\\
    \bottomrule
  \end{tabular}
  \caption{BLEU scores on meta-test split for different approaches evaluated across ten domains. Best results are
  highlighted in bold, results with-in two standard-deviations of the best value are underlined.}
  \label{tab:bleu:test}
\end{table*}

\begin{figure*}[h]
    \centering
    \includegraphics[width=\textwidth]{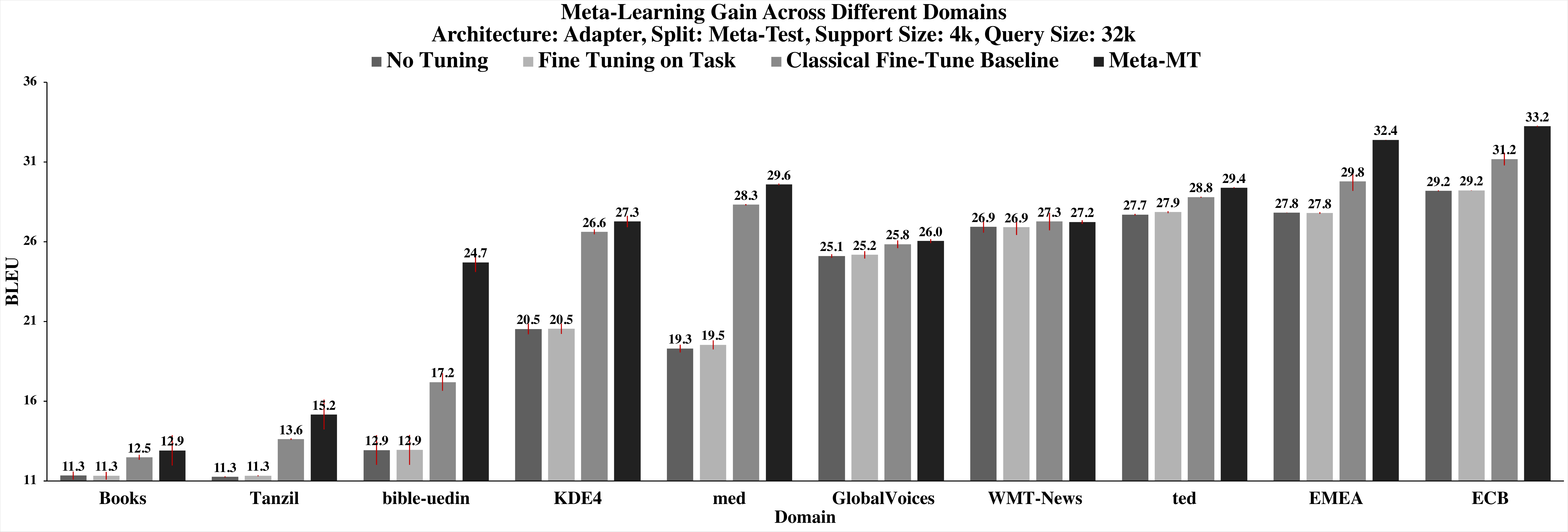}
    \caption{BLEU scores on meta-test split for different approaches evaluated across ten domains.}
    \label{fig:results}
\end{figure*}

\subsection{Evaluation Tasks and Metrics}

Our experimental setup operates as follows: using a collection of simulated machine translation adaptation tasks, we
train an NMT model $f_\theta$ using \ourname (\autoref{alg:train}). This model learns to adapt faster to new domains, by
fine-tuning on a tiny support set. Once $f_{\theta}$ is learned and fixed, we follow the test-time behavior described in~\autoref{sec:test}. We evaluate \ourname on the collection of ten different domains described in~\autoref{sec:data}.  We simulate domain adaptation problems by sub-sampling tasks with 4k English tokens for the support set, and 32k tokens for the query set.
We study the effect of varying the size of the query and the support sets in~\autoref{sec:size}. We use $N=160$ tasks for
the meta-training dataset $\cal D_{\textrm{meta-train}}$, where we sample $16$ tasks from each of the ten different
domains.  We use a meta-validation $\cal D_{\textrm{meta-test}}$ and meta-test $\cal D_{\textrm{meta-test}}$ sets of size
$10$, where we sample a single task from each domain. We report the mean and standard-deviation over three different
meta-test sets. For evaluation, we use BLEU~\citep{papineni2002bleu}.  We measure case-sensitive de-tokenized BLEU with
SacreBLEU~\citep{post2018call}. All results use beam search with a beam of size five. 

\subsection{Experimental Results} \label{sec:results}

Here, we describe our experimental results comparing the several algorithms from~\autoref{sec:baselines}. The overall
results are shown in~\autoref{tab:bleu:test} and~\autoref{fig:results}. \autoref{tab:bleu:test} shows the BLEU scores on
the meta-test dataset for all the different approaches across the ten domains. From these results we draw the following conclusions:
 \begin{enumerate}[noitemsep,nolistsep]
     \item The pre-trained En-De NMT model performs well on general domains. For instance, BLEU for WMT-News \footnote{This is subset of the full test set to match the sizes of query sets from other domains}, GlobalVoices, 
     and ECB is at least $26$ points. However, performance degrades on closed domains like Books, Quran, and Bible.
     [Column A].
     \item Domain adaptation by fine-tuning on a single task doesn't improve the BLEU score. This is expected, since
     we're only fine-tuning on 4k tokens (i.e. $\sim 200-300$ sentences) [A vs B].
     \item Significant leverage is gained by increasing the amount of fine-tuning data. Fine-tuning on all the available
     data used for meta-learning improves the BLEU score significantly across all domains. [B vs C]. To put this into
     perspective, this setup is tuned on all data aggregated  from all tasks: $160*4k$ words which is approximately
     $40K$ sentences.
     \item \ourname outperforms alternative domain adaptation approaches on all domains with negligible degradation on the baseline domain. \ourname is better than the
     non-adaptive baseline [A vs D], and succeeds in learning to adapt faster given the same amount of fine-tuning data
     [B vs D, C vs D]. Both \textbf{Fine-tuning on meta-train} [C] and \textbf{\ourname} [D] have access to exactly the same
     amount of training data, and both use the same model architecture. The difference however is in the learning
     algorithm. \ourname uses MAML (\autoref{alg:train}) to optimize the meta-objective function in~\autoref{eq:loss}.
     This ensures that the learned model initialization can easily be fine-tuned to new domains with very few examples.
 \end{enumerate}

\begin{figure}[bt!]
    \centering
    \includegraphics[scale=1.0,width=0.5\textwidth]{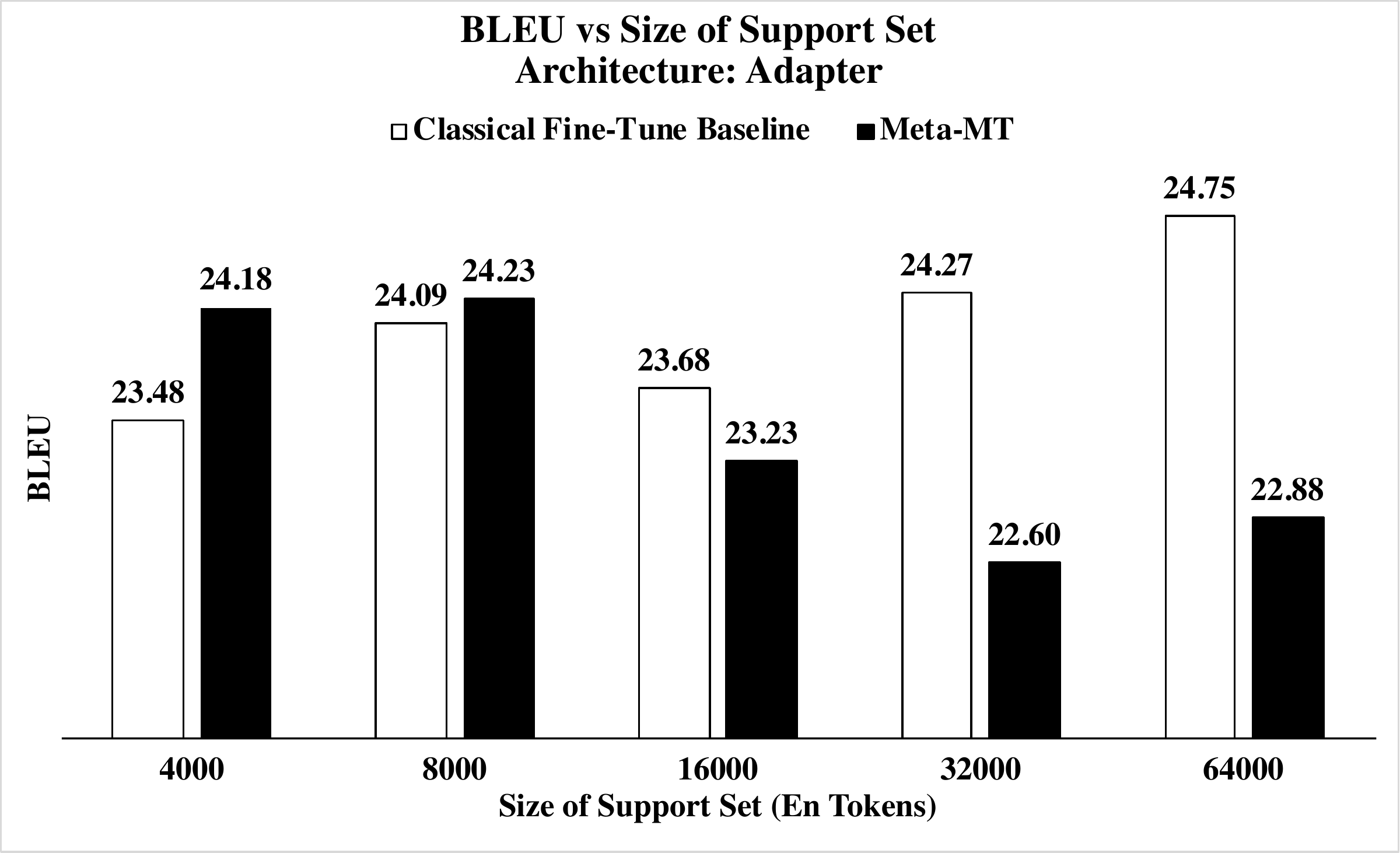}
    \caption{\ourname and fine-tuning adaptation performance on the meta-test set $\cal D_{\textrm{meta-test}}$ vs
    different support set sizes per adaptation task.}
    \label{fig:support}
\end{figure}

\begin{figure}
    \centering
    \includegraphics[scale=1.0,width=0.5\textwidth]{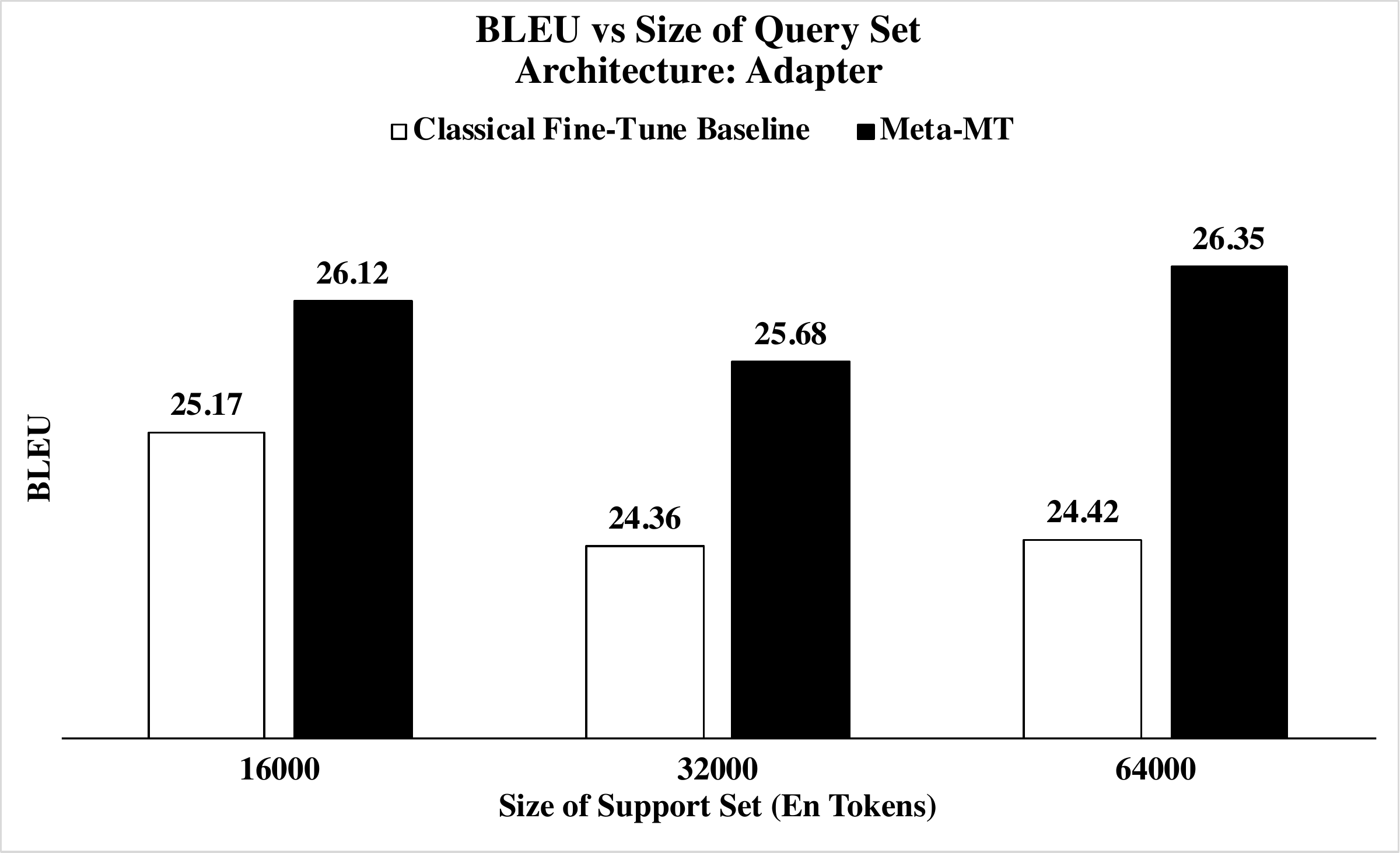}
    \caption{\ourname and fine-tuning adaptation performance on the meta-test set $\cal D_{\textrm{meta-test}}$ vs
    different query set sizes per adaptation task.}
    \label{fig:query}
\end{figure}

\subsection{Impact of Adaptation Task Size}
\label{sec:size}

To evaluate the effectiveness of \ourname when adapting with small in-domain corpora, we further compare the performance
of \ourname with classical fine-tuning on varying amounts of training data per adaptation task.  In~\autoref{fig:support}
we plot the overall adaptation performance on the ten domains when using different data sizes for the support set. In
this experiment, the only parameter that varies is the size of the task support set $\Tau_{\textrm{support}}$. We fix
the size of the query set per task to $16k$ tokens, and we vary the size of the support set from $4k$ to $64k$. To
ensure that the total amount of meta-training data $\cal D_{\textrm{meta-train}}$ is the same, we use $N=160$ tasks for
meta-training when the support size $\Tau_{\textrm{support}}$ is $4k$, $N=80$ tasks when the support size is $8k$,
$N=40$ tasks for support size of $16k$, $N=20$ tasks when the support size is $32k$, and finally $N=10$ meta-training
tasks when the support size is $64k$.  This controlled setup ensures that no setting has any advantage  by getting
access to additional amounts of training data. We notice that for reasonably small size of the support set ($4k$ and
$8k$), \ourname outperforms the classical fine-tuning baseline. However, when the data size increase ($16k$ to $64$),
\ourname is outperformed by the fine-tuning baseline. This happens because for a larger support size, e.g. $64k$, we
only have access to $10$ meta-training tasks in $\cal D_{\textrm{meta-train}}$, this is not enough to generalize to new
unseen adaptation tasks, and \ourname over-fits to the training tasks from $\cal D_{\textrm{meta-train}}$, however, the
performance degrades and doesn't generalize to $\cal D_{\textrm{meta-test}}$.

To understand more directly the impact of the query set on \ourname's performance, in~\autoref{fig:query} we show
\ourname and fine-tuning adaptation performance on the meta-test set $\cal D_{\textrm{meta-test}}$ on varying sizes for
the query set. We fix the support size to $4k$ and vary the query set size from $16k$ to $64k$. We observe that the edge
of improvement of \ourname over fine-tuning adaptation increases as we increase the size of the query set. For instance,
when we use a query set of size $64k$, \ourname outperforms fine-tuning by $1.93$ BLEU points, while the improvement is
only $0.95$ points when the query set is $16k$.

\subsection{Impact of Model Architecture}
\label{sec:architecture}

\begin{figure}
    \centering
    \includegraphics[scale=0.75, width=0.5\textwidth]{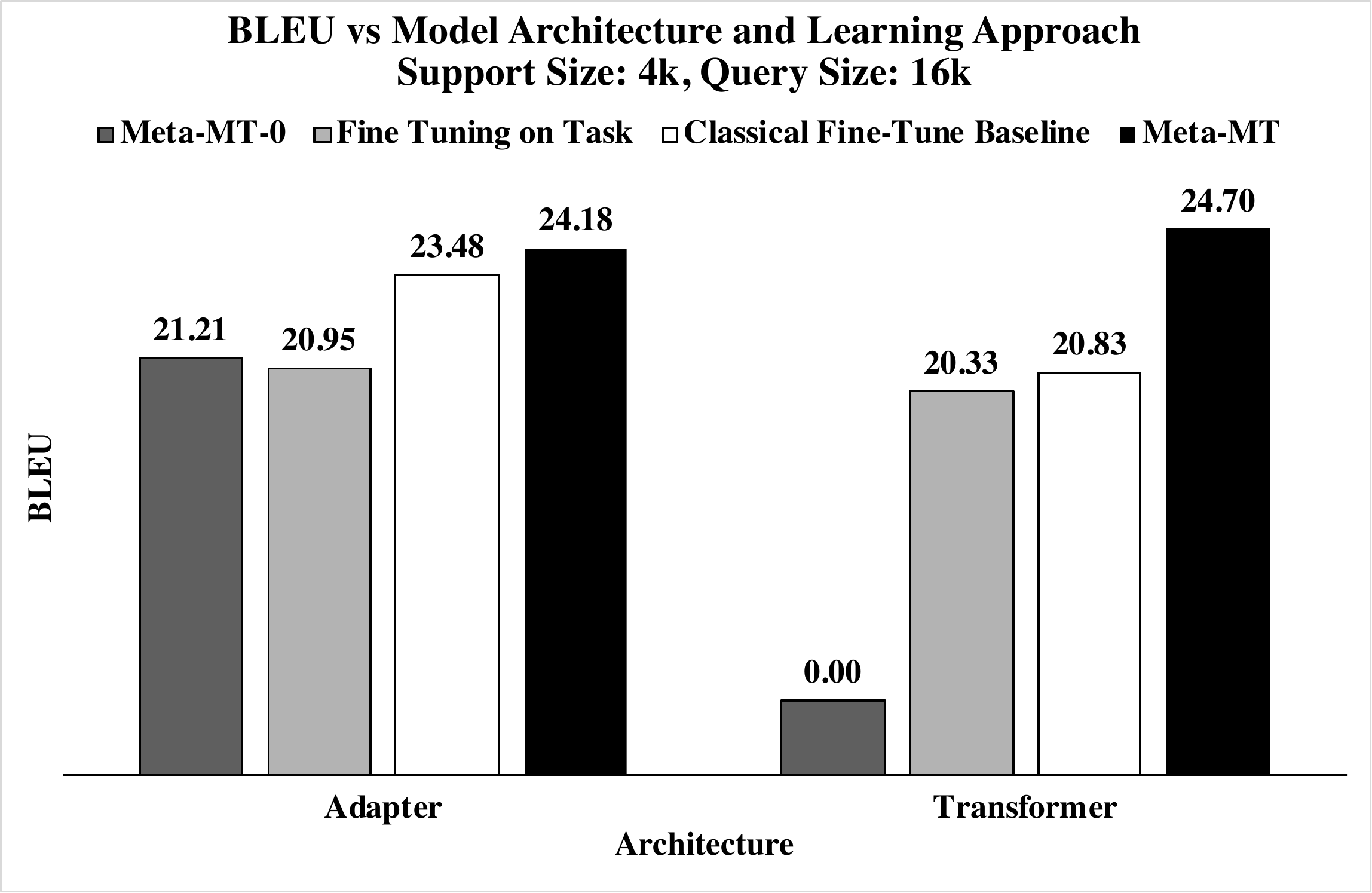}
    \caption{BLEU scores reported for two different model architectures: Adapter Transformer~\cite{bapna2019simple} (Left), and the Transformer base architecture~\cite{vaswani-etal-2012-smaller} (Right).}
    \label{fig:arch}
\end{figure}

In our experiments, we used the Adapter Transformer architecture~\cite{bapna2019simple}. This architecture fixes the
parameters of the pre-trained Transformer model, and only adapts the feed-forward adapter module. Our model included
$\sim 66M$ parameters, out of which we adapt only $405K$ (only $0.6\%$).  We found this adaptation strategy to be more
robust to meta-learning. To better understand this,~\autoref{fig:arch} shows the BLEU scores for the two different model
architectures. We find that while the meta-learned Transformer architecture (Right) slightly outperforms the Adapter
model (Left), it suffers from catastrophic forgetting: \textbf{\ourname-0} shows the zero-shot BLEU score before
fine-tuning the task on the support set. For the Transformer model, the score drops to zero and then quickly improves
once the parameters are tuned on the support set. This is undesirable, since it hurts the performance of the pre-trained
model, even on the general domain data. We notice that the Adapter model doesn't suffer from this problem.

\section{Conclusion}

We presented \ourname, a meta-learning approach for few shot NMT adaptation. We formulated few shot NMT adaptation  as a
meta-learning problem, and presented a strategy that learns better parameters for NMT systems that can be easily adapted
to new domains. We validated the superiority of \ourname to alternative domain adaptation approaches.  \ourname
outperforms alternative strategies in most domains using only a small fraction of fine-tuning data.

\section*{Acknowledgements}

The authors would like to thank members of the Microsoft Machine Translation Team as well as members of the
Computational Linguistics and Information Processing (CLIP) lab for reviewing earlier versions of this work.  Part of
this work was conducted when the first author was on a summer internship with Microsoft Research.  This material is
based upon work supported by the National Science Foundation under Grant No. 1618193. Any opinions, findings, and
conclusions or recommendations expressed in this material are those of the author(s) and do not necessarily reflect the
views of the National Science Foundation.

\bibliography{acl2020}
\bibliographystyle{acl_natbib}

\clearpage
\appendix
\begin{centering}
\Large
Supplementary Material For:\\
Meta-Learning for Few-Shot NMT Adaptation\\
\end{centering}
\makeappendix

\end{document}